\documentclass[journal]{journal}
\usepackage{tikz}
\usetikzlibrary{arrows.meta, fit, backgrounds, decorations.pathreplacing}
\usepackage{amsmath} 
\usepackage{booktabs}
\usepackage{multirow}
\usepackage{graphicx}
\usepackage{amsfonts}  
\usepackage{color}
\usepackage{array}   
\usepackage[ruled,vlined,linesnumbered]{algorithm2e}
\SetKwInOut{Require}{Require}

\begin{document}

\title{Comprehensive Comparisons of Reinforcement Learning Algorithms in Autonomous Driving}

\author{Dechen Xie and Mushuang Liu
\thanks{Dechen Xie and Mushuang Liu are with the Department of Mechanical Engineering, Virginia Polytechnic Institute and State University, Blacksburg, VA 24060, USA.{\tt\small dechenx@vt.edu, mushuang@vt.edu}}%
}

\maketitle
\thispagestyle{empty}

\begin{abstract}
This paper presents a comprehensive comparison of four representative reinforcement learning (RL) methods in autonomous vehicle (AV) decision-making, including: Fitted Value Iteration (FVI), Rollout Classification Policy Iteration (RCPI), Deep Q-Network (DQN), and a discrete-action adaptation of Deep Deterministic Policy Gradient (D-DDPG). The four methods are applied to the same AV decision-making problem at an unsignalized intersection and are compared in terms of optimality and safety, training stability, and training and decision cost. RCPI is the closest to a reference solution and has the most stable training process. However, it is the least data-efficient. D-DDPG reaches the second-smallest gap to a reference solution with the highest data efficiency, but has the least training stability. DQN is the most stable model-free method but has the highest collision rate. FVI achieves a lower collision rate than DQN, but is less data-efficient and less stable. Based on these results, we provide suggestions for selecting a method according to whether a model is available, the budget for training, and whether stable training is required.
\end{abstract}


\IEEEpeerreviewmaketitle

\section{Introduction}

\IEEEPARstart{A}{utonomous} Driving (AD) can bring benefits to society by improving road safety and enhancing transportation efficiency~\cite{FAGNANT2015167}. However, realizing these benefits requires Autonomous Vehicles (AVs) to make safe and efficient decisions in complex and dynamic traffic environments. Such decision-making remains a critical challenge~\cite{schwarting2018planning}.

Reinforcement Learning (RL) provides a mathematical framework for sequential decision-making, in which a policy is learned to maximize the cumulative reward over time~\cite{suttonbarto2018}. Since each driving decision affects the future traffic situation, RL has been widely explored for AV decision-making~\cite{zhu2022survey}. Most RL methods can be distinguished along two dimensions: whether a method uses a model of the environment and whether a method represents the policy explicitly.

A model of the environment predicts the resulting next state and reward from taking an action at a given state~\cite{suttonbarto2018}. Model-based methods can query the model at any state repeatedly to predict the results of taking an action without executing it. However, in AD, an exact model is not always available. In practice, the exact model is replaced by an approximate one which is either designed from prior knowledge or learned from collected data~\cite{moerland2022}. For example, Approximate Dynamic Programming (ADP) can be applied to solve path-tracking control problems with a designed environment model~\cite{LinADP}. Think2Drive learns a model of the driving environment from collected data and uses it to train a driving policy~\cite{li2024think2drive}. In contrast, model-free methods learn without a model. Instead, they execute an action in the environment and observe the resulting next state and reward. This procedure is closer to the way vehicles act in the real world. Deep Deterministic Policy Gradient (DDPG) has been applied to learn lane following on a real vehicle~\cite{kendall2019learning}. During training, the vehicle drove along the road under the current policy until the safety driver intervened. Every time step was recorded and used to update the learned policy. Deep Q-Network (DQN) has been used to learn how to cross unsignalized intersections in a traffic simulator~\cite{isele2018navigating}.

These methods can also be distinguished by whether they represent the policy explicitly. Value-based methods learn the value of states or actions and derive the policy from it. For example, DQN learns the value of each action and selects the action with the largest value~\cite{isele2018navigating}. Policy-based methods instead represent the policy explicitly. The ADP used in~\cite{LinADP} follows a policy iteration framework and learns a policy network. Actor-critic methods use an actor to learn a policy and a critic to learn the value of states or actions. Some studies treat actor-critic methods as a third category that combines value-based and policy-based methods~\cite{kiran2021}. Since the critic is only used to help the actor learn and actions are selected by the actor alone, we classify actor-critic methods as policy-based in this study. The DDPG agent for lane following described above is such an actor-critic method~\cite{kendall2019learning}.

Although many RL methods have been developed and implemented, they are rarely compared under identical conditions, and their respective strengths and limitations for AV decision-making remain unclear. A deeper understanding of these methods would benefit the selection and design of RL methods. This paper presents a comprehensive comparison of four RL methods, one for each combination of the two dimensions: Fitted Value Iteration (FVI), Rollout Classification Policy Iteration (RCPI), DQN, and a discrete-action adaptation of DDPG, denoted D-DDPG. FVI and RCPI are model-based, whereas DQN and D-DDPG are model-free; FVI and DQN are value-based, whereas RCPI and D-DDPG are policy-based, as summarized in Table~\ref{tab:classification}. The four methods are applied to the same decision-making problem at an unsignalized intersection and compared in terms of optimality and safety, training stability, and training and decision cost. Based on the results and features of each method, we provide suggestions for method selection.

\begin{table}[t]
\centering
\caption{Classification of the four RL methods compared in this paper along the two dimensions.}
\label{tab:classification}
\footnotesize
\renewcommand{\arraystretch}{1.15}
\begin{tabular}{@{}lcc@{}}
\toprule
& \textbf{Value-based} & \textbf{Policy-based} \\
\midrule
\textbf{Model-based} & FVI~\cite{munos2008finite} & RCPI~\cite{lagoudak} \\
\textbf{Model-free}  & DQN~\cite{mnih2015}        & D-DDPG~\cite{lillicrap2015,tilbury2023revisiting} \\
\bottomrule
\end{tabular}
\end{table}

The remainder of this paper is organized as follows. Section~\ref{sec:II} formulates the problem as a Deterministic Markov Decision Process (DMDP). Section~\ref{sec:preliminaries} introduces the preliminaries of RL. Section~\ref{sec:Methods} presents the four methods and how each of them computes its training target. Section~\ref{sec:numerical} presents the numerical results and suggestions for method selection. Section~\ref{sec:conclu} concludes the paper.

\section{Problem Formulation}
\label{sec:II}
In this section, we formulate the AV decision-making problem as a DMDP.

Consider a general traffic scenario with $N$ vehicles, where each vehicle is indexed by $i\in\{1,2,\ldots, N\}$. Let vehicle~1 be the controlled vehicle and vehicles $2,\ldots,N$ be the surrounding vehicles. The state of vehicle $i$ is denoted by $\xi_i\in\mathbb{R}^{n}$, where $n$ is the dimension of a single-vehicle state. In autonomous driving, the state of a vehicle often contains its position, velocity, and heading~\cite{kong2015}. The joint state is defined as $s=(\xi_1^\top,\ldots,\xi_N^\top)^\top\in\mathcal{S}$, where $\mathcal{S}\subseteq\mathbb{R}^{Nn}$ is the joint state space. The controlled vehicle follows a deterministic policy $\pi:\mathcal{S}\rightarrow\mathcal{A}$, where $\mathcal{A}$ is the action space and $|\mathcal{A}|$ denotes the number of actions in $\mathcal{A}$. In autonomous driving, the action
often contains acceleration and steering angle~\cite{kong2015}. In this work, the state space $\mathcal{S}$ is continuous and the action space $\mathcal{A}$ is finite. Given a state $s$, the controlled vehicle selects an action $a=\pi(s)\in\mathcal{A}$. The next state is determined by $s'=f(s,a)$, where $f:\mathcal{S}\times\mathcal{A}\rightarrow\mathcal{S}$ is the deterministic state-transition function. The controlled vehicle then receives an immediate reward $r(s,a)\in\mathbb{R}$, where $r:\mathcal{S}\times\mathcal{A}\rightarrow\mathbb{R}$ is the reward function. 

The decision-making process proceeds over discrete time steps $t\in\{0,1,\ldots\}$. At each time step $t$, the controlled vehicle observes the current state $s_t$, selects an action $a_t=\pi(s_t)$, transitions to the next state $s_{t+1}=f(s_t,a_t)$, and receives the reward $r(s_t,a_t)$. The observed sequence $\{s_0,a_0,r(s_0,a_0),s_1,a_1,\dots\}$ is called a trajectory. The state-value function $V^{\pi}(s)$ is the cumulative discounted reward collected along the trajectory that starts at $s$ and follows $\pi$,
\begin{equation}
V^{\pi}(s)=\sum_{t=0}^{\infty}\gamma^{t}\,r(s_t,a_t),\quad s_0=s,
\label{eq:state-value}
\end{equation}
where $\gamma\in[0,1)$ is the discount factor.
The objective is to find an optimal policy
$\pi^{*}$ that maximizes $V^{\pi}(s)$ at every state,
\begin{equation}
\pi^{*}=\arg\max_{\pi}V^{\pi}(s),\qquad \forall\,s\in\mathcal{S}.
\label{eq:optimal_policy}
\end{equation}

Since $s_{t+1}$ is determined by $s_t$ and $a_t$ alone, the Markov property holds. The problem can be modeled as a DMDP represented by the tuple
\begin{equation}
\mathcal{M}=\bigl(\mathcal{S},\mathcal{A},f,r,\gamma\bigr).
\end{equation}

\section{Preliminaries of Reinforcement Learning}
\label{sec:preliminaries}
 Solving \eqref{eq:optimal_policy} requires evaluating the cumulative reward of candidate policies, which depends on the transition function $f$ and the reward function $r$. An exact model is not always available in AD, since the driving policies of the surrounding vehicles are unknown. Instead of solving \eqref{eq:optimal_policy} with $f$ and $r$, RL starts from an estimate of the value function or the policy and corrects it with the reward and next state that result from taking an action in a state. This information can be obtained by querying either an exact or approximate model or by executing the action in the environment and observing the outcome. This section presents a brief introduction to RL.

To improve the estimate of the value function or the policy, an RL method needs to know how good it is to take an action at a state. RL measures this by the action-value function. For a policy $\pi$ and a state-action pair $(s,a)$, the action-value function $Q^{\pi}:\mathcal{S}\times\mathcal{A}\rightarrow\mathbb{R}$ is defined as
\begin{align}
Q^{\pi}(s,a)&=\sum_{t=0}^{\infty}\gamma^{t}\,r(s_t,a_t)
  \label{eq:q_def}\\
&=r(s,a)+\gamma\sum_{t=0}^{\infty}\gamma^{t}\,r(s_{t+1},a_{t+1})
  \nonumber\\
&=r(s,a)+\gamma\,Q^{\pi}\bigl(f(s,a),\,\pi(f(s,a))\bigr),
  \label{eq:q_pi}
\end{align}
where $s_{t+1}=f(s_t,a_t)$, $(s_0,a_0)=(s,a)$ and $a_t=\pi(s_t)$ for $t\geq 1$. Similarly, $V^{\pi}(s)$ satisfies
\begin{equation}
V^{\pi}(s)=r(s,\pi(s))+\gamma\,V^{\pi}\bigl(f(s,\pi(s))\bigr).
\label{eq:v_pi}
\end{equation}
Most policy-based methods follow two steps: policy evaluation and policy improvement. In policy evaluation, the current policy is evaluated by $V^{\pi}$ or $Q^{\pi}$. Policy improvement is conducted based on this evaluation by updating the policy toward actions with higher $V^{\pi}$ or $Q^{\pi}$.

The optimal value functions $V^{*}(s)=\max_{\pi}V^{\pi}(s)$ and $Q^{*}(s,a)=\max_{\pi}Q^{\pi}(s,a)$ satisfy the Bellman optimality equations~\cite{suttonbarto2018}
\begin{align}
V^{*}(s)&=\max_{a\in\mathcal{A}}\bigl[r(s,a)+\gamma\,V^{*}\bigl(f(s,a)\bigr)\bigr],
  \label{eq:v_opt}\\
Q^{*}(s,a)&=r(s,a)+\gamma\max_{a'\in\mathcal{A}}Q^{*}\bigl(f(s,a),a'\bigr).
  \label{eq:q_opt}
\end{align}
An optimal policy can be obtained from either function,
\begin{equation}
\begin{split}
\pi^{*}(s)&=\arg\max_{a\in\mathcal{A}}\bigl[r(s,a)+\gamma\,V^{*}\bigl(f(s,a)\bigr)\bigr]\\
&=\arg\max_{a\in\mathcal{A}}Q^{*}(s,a).
\end{split}
\label{eq:pi_opt}
\end{equation}

Value-based methods learn an estimate of the optimal value function, $V^{*}$ or $Q^{*}$, and derive the policy from it by \eqref{eq:pi_opt}. 

Both value-based and policy-based methods start from an initial estimate of the value function or the policy and improve it iteratively. Each update corrects the estimates with the actual reward and next state that result from taking an action at a state, which enter the update through the right-hand side of \eqref{eq:q_pi}, \eqref{eq:v_pi}, \eqref{eq:v_opt}, or \eqref{eq:q_opt}. We call this information an experience and represent it by the tuple $(s,a,r,s')$, where $r=r(s,a)$ and $s'=f(s,a)$.

An experience can be obtained in two ways. Given a model, the experience of any action at any state can be obtained by querying the model. Without a model, an experience is collected by executing an action in the environment, and only the outcome of the executed action is observed. Methods that rely on the first way are model-based, and methods that only need the second way are model-free~\cite{suttonbarto2018}. Since $f$ is deterministic, querying $f$ and $r$ at $(s,a)$ yields exactly the experience that executing $a$ at $s$ would produce. The data used by both kinds of methods can therefore be compared by the number of experiences.

\section{Reinforcement Learning Methods}
\label{sec:Methods}
In this section, we introduce the four RL methods compared in this paper in the order of FVI, DQN, RCPI, and D-DDPG. All four methods use neural networks as function approximators, so we first state the notation for them and then introduce each method in detail.

Throughout this paper, a function approximated by a neural network is written with its parameter as a subscript, e.g., $V_{\omega}$ denotes a network with parameter $\omega$ that represents a state-value function $V$, and $V_{\omega}(s)$ denotes its output at state $s$. The same convention is used for every network introduced below. Each network has its own parameter: $\omega$ for FVI, $\psi$ for DQN, $\eta$ for RCPI, and $\theta$ and $\phi$ for the actor and the critic of D-DDPG.

\subsection{FVI}
\label{sec:fvi}
FVI~\cite{munos2008finite} is a value-based and model-based method that learns an estimate of $V^{*}$. It extends Value Iteration~\cite{suttonbarto2018} by using a Neural Network (NN) $V_{\omega}$ to represent the state-value function. It maintains two networks: the online network $V_{\omega}$, which changes at every iteration, and a target network $V_{\omega^-}$, which is used to compute the targets. The two networks share the same architecture. $\omega^-$ is a delayed copy of $\omega$ which is updated by $\omega^-\leftarrow\omega$ every $M$ iterations. At iteration $k$ (for $k=1,\ldots,K$), FVI samples a set of states $\mathcal{D}\subset\mathcal{S}$ from a replay buffer $\mathcal{R}$. For each state $s\in\mathcal{D}$, it collects the experiences $(s,a,r,s')$ of all actions $a\in\mathcal{A}$ by querying the model. The value network is updated by minimizing the loss
\begin{equation}
\mathcal{L}_{V}^{(k)}(\omega)=\frac{1}{|\mathcal{D}|}\sum_{s\in\mathcal{D}}\bigl[V_{\omega}(s)-V_{\mathrm{target}}^k(s)\bigr]^2,
\label{eq:fvi_loss}
\end{equation}
yielding $\omega_k$, where $V_{\mathrm{target}}^k(s)$ is the target at iteration $k$ defined as
\begin{equation}
V_{\mathrm{target}}^k(s)=\max_{a\in\mathcal{A}}\bigl[r(s,a)+\gamma\,V_{\omega^-}\bigl(f(s,a)\bigr)\bigr].
\label{eq:ov_fvi_target}
\end{equation}
After $K$ iterations, the learned policy is derived as
\begin{equation}
\pi(s)=\arg\max_{a\in\mathcal{A}}\bigl[r(s,a)+\gamma\,V_{\omega_{K}}\bigl(f(s,a)\bigr)\bigr].
\label{eq:fvi_policy}
\end{equation}

Computing the target \eqref{eq:ov_fvi_target} requires solving an optimization problem over $a$ at the state $s$. The objective $r(s,a)+\gamma\,V_{\omega^-}(f(s,a))$ has to be evaluated over the action space, which requires knowing how $r(s,a)$ and $f(s,a)$ vary with $a$. FVI is therefore a model-based method. In this study, $\mathcal{A}$ is finite, so the maximum can be obtained by enumerating every $a\in\mathcal{A}$. Each evaluation requires one experience, so FVI requires $|\mathcal{A}|$ experiences to compute each target. The policy~\eqref{eq:fvi_policy} also takes the maximum over all actions at the current state, so FVI requires the model not only in training but also at every decision.

The FVI training procedure is summarized in Algorithm~\ref{alg:fvi}.
\begin{algorithm}[t]
\caption{Fitted Value Iteration (FVI)}
\label{alg:fvi}
\Require{access to the model of the environment, discount factor $\gamma$, target update period $M$}
Initialize the online network $V_{\omega_0}$ and the target network $V_{\omega^-}$ with $\omega^-\leftarrow\omega_0$\;
Initialize the replay buffer $\mathcal{R}$\;
\For{iteration $k=1,\ldots,K$}{
    Sample a set of states $\mathcal{D}$ from $\mathcal{R}$\;
    Collect the experiences $(s,a,r,s')$ of all actions $a\in\mathcal{A}$ for each $s\in\mathcal{D}$ by querying the model\;
    Compute the target $V_{\mathrm{target}}^k(s)$ for each $s\in\mathcal{D}$ according to \eqref{eq:ov_fvi_target}\;
    Update the online network by minimizing the loss~\eqref{eq:fvi_loss}, yielding $\omega_k$\;
    Every $M$ iterations, set $\omega^-\leftarrow\omega_k$\;
}
\Return the value network $V_{\omega_K}$\;
\end{algorithm}

\subsection{DQN}
\label{sec:dqn}
DQN~\cite{mnih2015} is a value-based and model-free method. It extends Q-learning~\cite{suttonbarto2018} with function approximation. Instead of learning the value of a state, DQN uses an NN $Q_{\psi}$ to represent the action-value function, which outputs $Q_{\psi}(s,a)$ for every $a\in\mathcal{A}$ at a given state $s$. DQN also maintains an online network $Q_{\psi}$ and a target network $Q_{\psi^-}$, where $\psi^-$ is updated by $\psi^-\leftarrow\psi$ every $M$ iterations. Experiences are collected by interacting with the environment under an $\epsilon$-greedy policy and stored in a replay buffer $\mathcal{R}$. At state $s$, a random action is selected with a probability of $\epsilon$, and otherwise, the action $\arg\max_{a\in\mathcal{A}}Q_{\psi}(s,a)$ is selected. At iteration $k$ (for $k=1,\ldots,K$), DQN executes one action in the environment, stores the resulting experience in $\mathcal{R}$, and samples a batch of experiences $\mathcal{B}$ from $\mathcal{R}$. The online network is updated by minimizing the loss
\begin{equation}
\mathcal{L}_{Q}(\psi)=\frac{1}{|\mathcal{B}|}\sum_{(s,a,r,s')\in\mathcal{B}}
\bigl[Q_{\psi}(s,a)-Q_{\mathrm{target}}(s,a)\bigr]^2,
\label{eq:dqn_loss}
\end{equation}
yielding $\psi_k$, where $Q_{\mathrm{target}}(s,a)$ is the target defined as
\begin{equation}
Q_{\mathrm{target}}(s,a)=r+\gamma\max_{a'\in\mathcal{A}}Q_{\psi^-}(s',a').
\label{eq:ov_dqn_target}
\end{equation}
After $K$ iterations, the learned policy is derived as
\begin{equation}
\pi(s)=\arg\max_{a\in\mathcal{A}}Q_{\psi_K}(s,a).
\label{eq:dqn_policy}
\end{equation}

Unlike \eqref{eq:ov_fvi_target}, the maximum in \eqref{eq:ov_dqn_target} is taken over the actions at the next state $s'$. The action-values $Q_{\psi^-}(s',a')$ for every $a'\in\mathcal{A}$ can be provided by the target network itself. Computing the target therefore only requires one experience $(s,a,r,s')$, which can be collected by executing $a$ at $s$ in the environment. DQN is thus a model-free method. 

The DQN training procedure is summarized in Algorithm~\ref{alg:dqn}.

\begin{algorithm}[t]
\caption{Deep Q-Network (DQN)}
\label{alg:dqn}
\Require{access to the environment for interaction, discount factor $\gamma$, exploration rate $\epsilon$, target update period $M$}
Initialize the online network $Q_{\psi_0}$ and the target network $Q_{\psi^-}$ with $\psi^-\leftarrow\psi_0$\;
Initialize the replay buffer $\mathcal{R}$\;
\For{iteration $k=1,\ldots,K$}{
    Observe $s$, select $a$ by the $\epsilon$-greedy policy with $Q_{\psi_{k-1}}$, execute it, and store $(s,a,r,s')$ in $\mathcal{R}$\;
    Sample a batch of experiences $\mathcal{B}$ from $\mathcal{R}$\;
    Compute the target $Q_{\mathrm{target}}(s,a)$ for each experience in $\mathcal{B}$ according to \eqref{eq:ov_dqn_target}\;
    Update the online network by minimizing the loss~\eqref{eq:dqn_loss}, yielding $\psi_k$\;
    Every $M$ iterations, set $\psi^-\leftarrow\psi_k$\;
}
\Return the action-value network $Q_{\psi_K}$\;
\end{algorithm}

\subsection{RCPI}
\label{sec:rcpi}
RCPI~\cite{lagoudak} is a policy-based and model-based method that learns an estimate of $\pi^{*}$. It extends Policy Iteration~\cite{suttonbarto2018} by using an NN $\pi_{\eta}$ to represent the policy. Given a state $s$, the policy network outputs a vector $\pi_{\eta}(s)\in\mathbb{R}^{|\mathcal{A}|}$, whose entry $\pi_{\eta}(s,a)$ is the score of action $a\in\mathcal{A}$. The policy selects the action with the highest score, $\pi(s)=\arg\max_{a\in\mathcal{A}}\pi_{\eta}(s,a)$. At iteration $k$ (for $k=1,\ldots,K$), RCPI samples a set of states $\mathcal{D}\subset\mathcal{S}$ from a replay buffer $\mathcal{R}$ and fixes the current policy as $\pi_k=\pi_{\eta_{k-1}}$. The policy network is updated to predict the target action $a_{\mathrm{target}}(s)$ by minimizing the loss
\begin{equation}
\mathcal{L}_{\pi}^{(k)}(\eta)=-\frac{1}{|\mathcal{D}|}\sum_{s\in\mathcal{D}}
\log\frac{\exp\bigl(\pi_{\eta}(s,a_{\mathrm{target}}(s))\bigr)}{\sum_{a\in\mathcal{A}}\exp\bigl(\pi_{\eta}(s,a)\bigr)},
\label{eq:rcpi_loss}
\end{equation}
yielding $\eta_k$, where the target action is defined as
\begin{equation}
a_{\mathrm{target}}(s)=\arg\max_{a\in\mathcal{A}}Q^{\pi_k}(s,a).
\label{eq:ov_rcpi_target}
\end{equation}
RCPI computes $Q^{\pi_k}(s,a)$ according to \eqref{eq:q_def}. For each action $a$ at state $s$, it collects the experiences along the trajectory that starts from $s$ with action $a$ and then follows $\pi_k$ by querying the model. RCPI is therefore a model-based method. In this study, $\mathcal{A}$ is finite, so RCPI simulates one trajectory for every $a\in\mathcal{A}$. Let $H$ be the number of time steps of a trajectory. Each step requires one experience, so RCPI requires $|\mathcal{A}|H$ experiences to compute each target. After $K$ iterations, the learned policy is $\pi(s)=\arg\max_{a\in\mathcal{A}}\pi_{\eta_K}(s,a)$. The learned policy is obtained from the network alone, so RCPI does not need the model at deployment. 

The RCPI training procedure is summarized in Algorithm~\ref{alg:rcpi}.
\begin{algorithm}[t]
\caption{Rollout Classification Policy Iteration (RCPI)}
\label{alg:rcpi}
\Require{access to the model of the environment, discount factor $\gamma$}
Initialize the policy network $\pi_{\eta_0}$ and the replay buffer $\mathcal{R}$\;
\For{iteration $k=1,\ldots,K$}{
    Sample a set of states $\mathcal{D}$ from $\mathcal{R}$\;
    Fix the current policy $\pi_k=\pi_{\eta_{k-1}}$\;
    For each $s\in\mathcal{D}$ and $a\in\mathcal{A}$, collect the experiences along the trajectory that starts from $s$ with action $a$ and then follows $\pi_k$ by querying the model\;
    Compute $Q^{\pi_k}(s,a)$ according to \eqref{eq:q_def}\;
    Compute the target action $a_{\mathrm{target}}(s)$ for each $s\in\mathcal{D}$ according to \eqref{eq:ov_rcpi_target}\;
    Update the policy network by minimizing the loss~\eqref{eq:rcpi_loss}, yielding $\eta_{k}$\;
}
\Return the policy network $\pi_{\eta_K}$\;
\end{algorithm}

\subsection{D-DDPG}
\label{sec:ddpg}
D-DDPG is a policy-based and model-free method. It is derived from DDPG~\cite{lillicrap2015} by adapting it to a discrete action space. It follows the actor-critic structure. Like RCPI, the actor uses a policy network $\pi_{\theta}$ to represent the policy, which outputs a score vector $\pi_{\theta}(s)$ whose entry $\pi_{\theta}(s,a)$ is the score of $a\in\mathcal{A}$. Instead of calculating $Q^{\pi}(s,a)$ from trajectories, D-DDPG introduces a critic network $Q_{\phi}$ to learn $Q^{\pi}(s,a)$.

Standard DDPG is designed for continuous action spaces. The actor is updated through the gradient of the critic, which requires the selected action to be differentiable with respect to the actor parameters. This differentiability does not hold in a discrete action space. To address this, we adopt the straight-through Gumbel-Softmax (STGS) estimator~\cite{tilbury2023revisiting}. STGS passes the gradient to the actor through a softmax vector over the action scores, so the critic has to take an action input that lies in the same space as this vector. We therefore encode each action $a\in\mathcal{A}$ as a one-hot vector $\mathbf{e}_a\in\{0,1\}^{|\mathcal{A}|}$, whose entry corresponding to $a$ is $1$ and all other entries are $0$. Every action is passed to the critic in this form. During training, independent Gumbel noise $g_a\sim\mathrm{Gumbel}(0,1)$ is added to each score. The action with the largest perturbed score,
\begin{equation}
  a^\star=\arg\max_{a\in\mathcal{A}}\bigl[\pi_{\theta}(s,a)+g_a\bigr],
\label{eq:stgs}
\end{equation}
is selected and passed to the critic as $\mathbf{e}_{a^\star}$. In the backward pass, the gradient of $\mathbf{e}_{a^\star}$ is replaced by that of the tempered softmax $\mathrm{softmax}\bigl((\pi_{\theta}(s)+g)/\tau_{\mathrm{GS}}\bigr)$, where $g$ is the vector of the noise $g_a$ and $\tau_{\mathrm{GS}}>0$ is the temperature. Therefore the gradient of the critic output with respect to $\theta$ can be propagated back to the actor.

D-DDPG maintains an online actor $\pi_{\theta}$ and an online critic $Q_{\phi}$, together with a target actor $\pi_{\theta^-}$ and a target critic $Q_{\phi^-}$. At each iteration, the target parameters are moved toward the online parameters by a soft update with coefficient $\tau_{\mathrm{target}}$, which stabilizes training~\cite{lillicrap2015}. At iteration $k$ (for $k=1,\ldots,K$), D-DDPG selects one action by \eqref{eq:stgs}, executes it in the environment, stores the resulting experience in a replay buffer $\mathcal{R}$, and samples a batch of experiences $\mathcal{B}$ from $\mathcal{R}$. The online critic is updated by minimizing the loss
\begin{equation}
\mathcal{L}_{\mathrm{critic}}(\phi)=\frac{1}{|\mathcal{B}|}\sum_{(s,a,r,s')\in\mathcal{B}}\bigl[Q_{\phi}(s,a)-Q_{\mathrm{target}}(s,a)\bigr]^2,
\label{eq:ddpg_critic_loss}
\end{equation}
yielding $\phi_k$, where $Q_{\mathrm{target}}(s,a)$ is the target defined as
\begin{equation}
Q_{\mathrm{target}}(s,a)=r+\gamma\,Q_{\phi^-}(s',a^{\pi}),
\label{eq:ov_ddpg_target}
\end{equation}
and $a^{\pi}$ is the action given by \eqref{eq:stgs} at $s'$ with the target actor $\pi_{\theta^-}$.

The online actor is then updated by minimizing the loss
\begin{equation}
\mathcal{L}_{\mathrm{actor}}(\theta)=-\frac{1}{|\mathcal{B}|}\sum_{s\in\mathcal{B}}Q_{\phi}\bigl(s,a^{\star}\bigr),
\label{eq:ddpg_actor_loss}
\end{equation}
yielding $\theta_k$, where $a^{\star}$ is given by the online actor $\pi_{\theta}$. After $K$ iterations, the learned policy is derived as
\begin{equation}
\pi(s)=\arg\max_{a\in\mathcal{A}}\pi_{\theta_K}(s,a).
\label{eq:ddpg_policy}
\end{equation}

In \eqref{eq:ov_ddpg_target}, the action at $s'$ is given by the target actor and its value by the target critic. Computing the target therefore only requires one experience $(s,a,r,s')$, which can be collected by executing $a$ at $s$ in the environment. D-DDPG is thus a model-free method and requires one experience to compute each target. The policy~\eqref{eq:ddpg_policy} is obtained from the actor alone, so D-DDPG does not need the model at deployment.

The D-DDPG training procedure is summarized in Algorithm~\ref{alg:discrete_ddpg}.

\begin{algorithm}[t]
\caption{Discrete DDPG (D-DDPG)}
\label{alg:discrete_ddpg}
\Require{access to the environment for interaction, discount factor $\gamma$, temperature $\tau_{\mathrm{GS}}$, soft-update coefficient $\tau_{\mathrm{target}}$}
Initialize the online actor $\pi_{\theta_0}$ and the online critic $Q_{\phi_0}$\;
Initialize the target actor $\pi_{\theta^-}$ and the target critic $Q_{\phi^-}$ with $\theta^-\leftarrow\theta_0$ and $\phi^-\leftarrow\phi_0$\;
Initialize the replay buffer $\mathcal{R}$\;
\For{iteration $k=1,\ldots,K$}{
    Observe $s$, select $a^{\star}$ by \eqref{eq:stgs} with $\pi_{\theta_{k-1}}$, execute it, and store $(s,a^{\star},r,s')$ in $\mathcal{R}$\;
    Sample a batch of experiences $\mathcal{B}$ from $\mathcal{R}$\;
    Compute the target $Q_{\mathrm{target}}(s,a)$ for each experience in $\mathcal{B}$ according to \eqref{eq:ov_ddpg_target}\;
    Update the online critic by minimizing the loss~\eqref{eq:ddpg_critic_loss}, yielding $\phi_k$\;
    Update the online actor by minimizing the loss~\eqref{eq:ddpg_actor_loss}, yielding $\theta_k$\;
    $\theta^-\leftarrow(1-\tau_{\mathrm{target}})\theta^-+\tau_{\mathrm{target}}\theta_k$\;
    $\phi^-\leftarrow(1-\tau_{\mathrm{target}})\phi^-+\tau_{\mathrm{target}}\phi_k$\;
}
\Return the policy network $\pi_{\theta_K}$\;
\end{algorithm}

\section{Numerical Results}
\label{sec:numerical}
To compare the four RL methods, we test them on a simulated unsignalized intersection scenario. Each method is evaluated in three aspects: optimality and safety, training stability, and training and decision cost. Optimality is measured by the gap in discounted reward between the learned policy and a reference solution. Safety is measured by the collision rate of the learned policy. Training stability is presented by the learning curve and is measured by the drop from the peak discounted reward to the discounted reward at the end of training. Training cost is measured by the number of experiences required to reach given levels of discounted reward. Decision cost is measured by the time the learned policy takes to make a decision at one state.

In the following subsections, we first introduce the simulation setup in Section~\ref{sec:SimulationSetup}, then compare the four methods in the three aspects in Section~\ref{sec:results}, and finally provide suggestions for method selection in Section~\ref{sec:selection}.

\subsection{Simulation Setup}
\label{sec:SimulationSetup}
\subsubsection{Environment Setup}
\begin{figure}[t]
  \centering
  \includegraphics[width=0.55\columnwidth]{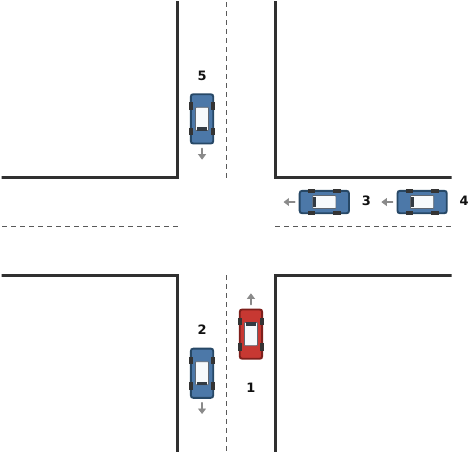}
  \caption{Illustration of the unsignalized intersection.}
  \label{fig:intersection}
\end{figure}

We consider an unsignalized intersection environment with $N=5$ vehicles as illustrated in Fig.~\ref{fig:intersection}. Vehicle~1 is the controlled vehicle and vehicles $2,\ldots,5$ are the surrounding vehicles that are moving at a constant speed. All vehicles travel straight through the intersection. The goal is to find the optimal policy \eqref{eq:optimal_policy} of the controlled vehicle, which maximizes the discounted reward~\eqref{eq:state-value} with discount factor $\gamma=0.99$.

The intersection is described in a fixed Cartesian frame with its origin at the intersection center. The $x$-axis points east and the $y$-axis points north. The state of vehicle~$i$ is
\begin{equation}
  \xi_{i}=[\,x_{i},\;y_{i},\;v_{x,i},\;v_{y,i}\,]^{\top},
  \label{eq:vehicle_state}
\end{equation}
where $(x_{i},y_{i})$ is its position and $(v_{x,i},v_{y,i})$ is its velocity. The joint state defined in Section~\ref{sec:II} is therefore
\begin{equation}
  s=(\xi_{1}^{\top},\xi_{2}^{\top},\ldots,\xi_{5}^{\top})^{\top}\in\mathbb{R}^{20}.
  \label{eq:joint_state}
\end{equation}

The environment proceeds with a time step $\Delta t=0.25$~s. A simulation terminates when the controlled vehicle passes the intersection and reaches $y_{1}>30$~m, when a collision occurs, or after $80$ time steps ($20$~s), whichever occurs first. A collision is detected when $d_{i}<0.5$~m for any $i\in\{2,\ldots,N\}$, where $d_{i}$ is the Euclidean distance between the controlled vehicle and surrounding vehicle $i$,
\begin{equation}
d_{i}=\sqrt{(x_{1}-x_{i})^{2}+(y_{1}-y_{i})^{2}}.
\label{eq:distance}
\end{equation}

Each vehicle is modeled as a point mass. The controlled vehicle travels along the $y$-axis in the positive direction. Its action $a$ is its longitudinal acceleration. The finite action space of the controlled vehicle is
\begin{equation}
\mathcal{A}=\{-4,-3,\ldots,3,4\}~\mathrm{m/s^{2}}.
\label{eq:actionspace}
\end{equation}
There are a total of nine actions in $\mathcal{A}$ so $|\mathcal{A}|=9$ in this work.
The transition function introduced in Section~\ref{sec:II} is therefore
\begin{equation}
  f(s,a)=\bigl((A\xi_{1}+Ba)^{\top},\,(A\xi_{2})^{\top},\ldots,
                (A\xi_{N})^{\top}\bigr)^{\top},
  \label{eq:transition}
\end{equation}
with
\begin{equation}
  A=\begin{bmatrix}
    1&0&\Delta t&0\\ 0&1&0&\Delta t\\ 0&0&1&0\\ 0&0&0&1
  \end{bmatrix},
  \qquad
  B=\begin{bmatrix} 0\\ 0\\ 0\\ \Delta t \end{bmatrix}.
  \label{eq:AB}
\end{equation}

Following the reward design in \cite{liu2026realtime}, the reward function $r$ is defined as
\begin{equation}
  r(s,a)=-\alpha\left(\frac{v'_{y,1}-v_{\mathrm{ref}}}{v_{\mathrm{ref}}}\right)^{2}
       -\beta\sum_{i=2}^{N}\frac{1}{(d'_{i})^{2}+\delta},
  \label{eq:reward_state}
\end{equation}
where $v'_{y,1}$ and $d'_{i}$ are the longitudinal velocity of the controlled vehicle and its distance to surrounding vehicle $i$ at the next state $s'=f(s,a)$. Here $v_{\mathrm{ref}}=5$~m/s is the desired speed, $\delta=0.01$ prevents division by zero, and $\alpha=6$ and $\beta=100$ are weighting coefficients. The first term of $r$ penalizes deviation from the desired speed, while the second term penalizes proximity to surrounding vehicles. 

\subsubsection{Training Setup}
\begin{figure*}[t]
    \centering
    \includegraphics[width=\textwidth]{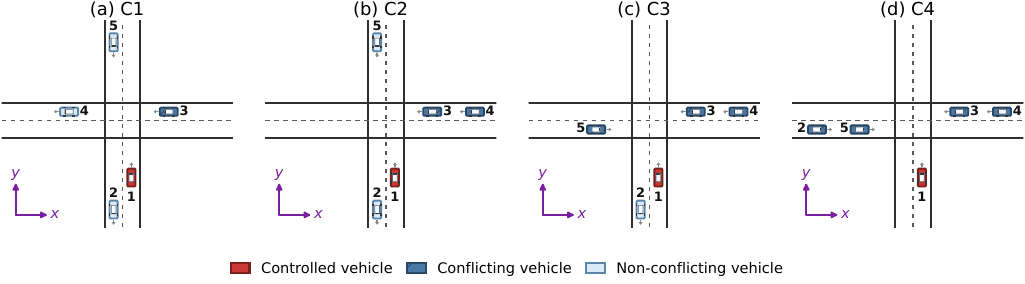}
    \caption{Intersection layouts C1--C4. Vehicle~1 (red) is the controlled vehicle. A surrounding vehicle is conflicting (dark blue) or non-conflicting (light blue).}
    \label{fig:intersection_scenarios}
\end{figure*}

In practice, different layouts of surrounding vehicles may occur within the same scenario. A method therefore needs to learn a general policy that works across multiple layouts. In this study, we design four layouts of the unsignalized intersection scenario, denoted C1--C4, as illustrated in Fig.~\ref{fig:intersection_scenarios}. Each method learns a single policy that is trained on all four layouts.
We characterize the complexity of a layout by the number of conflicting vehicles, i.e., surrounding vehicles whose future paths cross the path of the controlled vehicle. The positions and travel directions of the surrounding vehicles are assigned so that C1, C2, C3, and C4 contain 1, 2, 3, and 4 conflicting vehicles, respectively. Every additional conflicting vehicle decreases the number of feasible safe decisions of the controlled vehicle. The complexity of the layout thus increases from C1 to C4.

All four methods sample their training data from a replay buffer $\mathcal{R}$. The replay buffer of model-based methods FVI and RCPI stores states only. To cover the states that the controlled vehicle  can reach from the initial states, these states are taken from trajectories that start from an initial
state sampled from the four layouts and select the action at every step uniformly from $\mathcal{A}$ until termination. Every state on these trajectories is stored in $\mathcal{R}$. Model-free methods DQN and D-DDPG cannot query the model, so $\mathcal{R}$ stores experiences $(s,a,r,s')$ collected by executing actions in the environment. The experiences are collected by an $\epsilon$-greedy policy for DQN and by \eqref{eq:stgs} for D-DDPG. In all four methods, $\mathcal{R}$ is first-in-first-out, and $|\mathcal{D}|=64$ states or $|\mathcal{B}|=64$ experiences are sampled uniformly from it at each iteration.

\begin{figure}[t]
  \centering
  \includegraphics[width=\columnwidth]{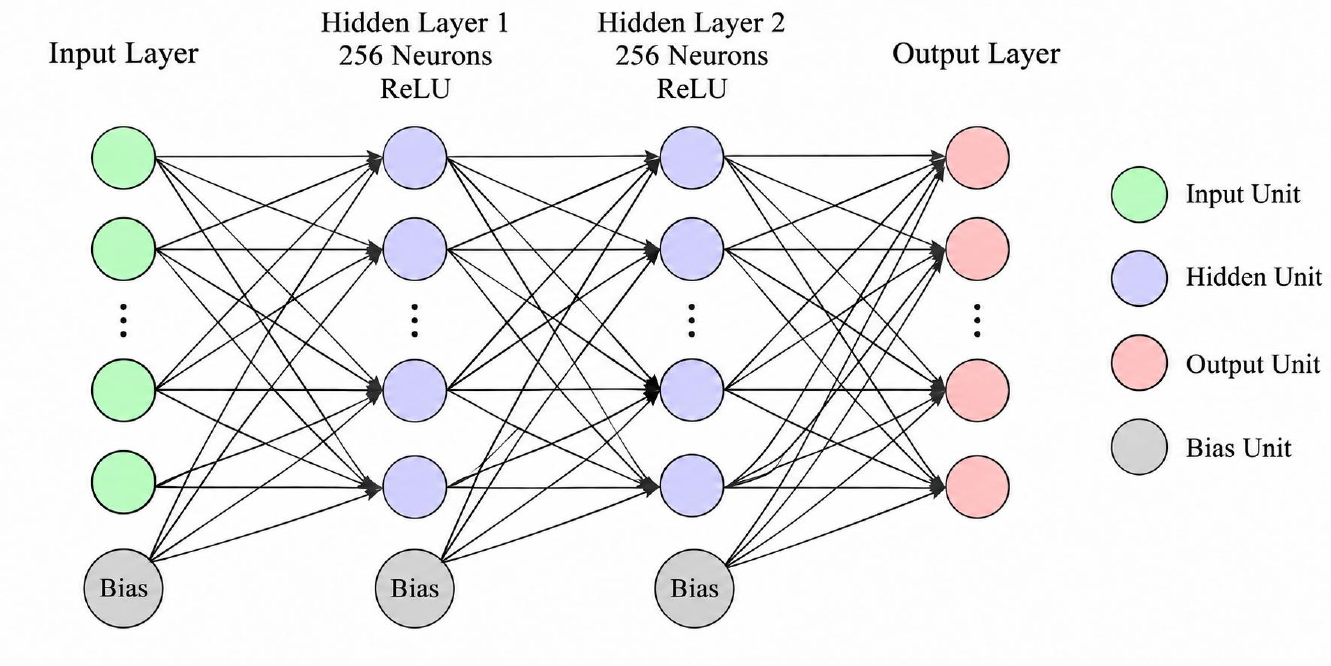}
  \caption{Shared network architecture.}
  \label{fig:architecture}
\end{figure}

All four methods use NNs as function approximators. As shown in Fig.~\ref{fig:architecture}, all networks share the same backbone of two fully connected hidden layers with 256 units each and ReLU activations with same initialization. The networks differ only in their input and output layers, which depend on the method. For FVI, DQN, RCPI, and the actor of D-DDPG, the input is the state $s$. The value network of FVI outputs the scalar $V_{\omega}(s)$, the Q-network of DQN outputs $Q_{\psi}(s,a)$ for every $a\in\mathcal{A}$, and the policy network of RCPI and the actor of D-DDPG output the scores $\pi_{\eta}(s,a)$ and $\pi_{\theta}(s,a)$, respectively, for every $a\in\mathcal{A}$. The critic of D-DDPG takes $s$ and an action $a$, encoded as a one-hot vector, as input and outputs the scalar $Q_{\phi}(s,a)$.

We measure training progress by the number of network updates. For FVI, RCPI, and DQN, one network update is one update of the parameters of $V_{\omega}$, $\pi_{\eta}$, and $Q_{\psi}$, respectively. For D-DDPG, one network update includes one update of the critic $Q_{\phi}$ and one update of the actor $\pi_{\theta}$. The number of network updates measures how much each method has learned. Each method is trained for $2.5\times 10^{6}$ network updates with five random seeds. This budget is chosen to make sure that every method is trained sufficiently.

\subsubsection{Evaluation Setup}
\label{sec:eval_setup}
To obtain the training outcome of each method, the policies at all recorded checkpoints are first evaluated on a validation set of 500 initial states that are sampled equally from each of the four layouts. For each method and each seed, the checkpoint with the highest mean discounted reward over the validation set is considered as the peak and selected as the learned policy. The learned policies are then evaluated on a separate test set of 4000 initial states, sampled in the same way as the validation set.

To better understand the optimality of each method, we obtain a reference solution at each initial state by searching over action sequences with Differential Evolution (DE). DE maximizes the discounted reward of an 80-step action sequence over the same action space. It uses a population
of $8{,}000$ and $500$ generations. Since DE is a heuristic optimizer, the reference is near-optimal rather than
exact, so a policy can occasionally outperform it in individual states. We evaluate the optimality of each method by the gap to the reference and the safety by the collision rate. The gap is defined as
\begin{equation}
  \mathrm{Gap}=\bar{V}^{\mathrm{DE}}-\bar{V}^{\pi},
  \label{eq:gap}
\end{equation}
where $\bar{V}^{\pi}$ and $\bar{V}^{\mathrm{DE}}$ are the mean discounted reward of the learned policy and of the DE reference, respectively. A smaller gap indicates a policy closer to the reference. The collision rate is the percentage of the trajectories of the learned policy that end in a collision. Collisions among the surrounding vehicles are not considered. For all 4500 sampled states, the DE solution is collision-free. 

We measure training stability by the learning curve and by the drop from the peak discounted reward to the discounted reward at the end of training. The learning curve shows the learning process through the evaluation of all recorded checkpoints on the validation set. The drop is calculated by
\begin{equation}
  \mathrm{Drop}=\frac{\bar{V}_{\mathrm{best}}-\bar{V}_{\mathrm{final}}}{\bigl|\bar{V}^{\mathrm{DE}}\bigr|},
  \label{eq:drop}
\end{equation}
where $\bar{V}_{\mathrm{best}}$ is the mean discounted reward of the selected checkpoint, $\bar{V}_{\mathrm{final}}$ is the mean discounted reward averaged over the last ten checkpoints, and $\bar{V}^{\mathrm{DE}}$ is the mean discounted reward of the DE reference. All three quantities are computed on the validation set. A large drop indicates that the peak reached during training is not retained.

We measure the training cost and decision cost of each method by the number of experiences it uses and by its decision time. The number of experiences is recorded when the mean discounted reward first exceeds $2\bar{V}^{\mathrm{DE}}$ and $1.5\bar{V}^{\mathrm{DE}}$, respectively, and when the peak is reached. Since the reward \eqref{eq:reward_state} is non-positive, $\bar{V}^{\mathrm{DE}}<0$. Therefore, $2\bar{V}^{\mathrm{DE}}$ is the lower threshold. An experience is counted each time it is used to compute a target defined in Section~\ref{sec:Methods}. The decision time is measured on a set of 100 states sampled equally from the four layouts. For each method, the selected checkpoint of every seed is first warmed up on 100 states. The decision time is then recorded for making the decision at each state, one state at a time. This is repeated five times. For DE, the decision time is the time required to optimize the action sequence from a given state. It is measured at each of the 100 states and repeated five times with different random seeds, using the same parameters as for the reference solution. We report the mean and the standard deviation of the time per decision.

\subsection{Comparison Results}
\label{sec:results}
\subsubsection{Optimality and Safety}
Table~\ref{tab:performance} reports the gap to the DE reference and the collision rate under the four layouts.

\begin{table*}[t]
\centering
\caption{Optimality gap and collision rate over four layouts. The smallest gap in each column is in bold. Col.\% is the collision rate in percent.}
\label{tab:performance}

\setlength{\tabcolsep}{6.0pt}
\renewcommand{\arraystretch}{1.10}
\small

\begin{tabular}{lcccccccccc}
\toprule
& \multicolumn{2}{c}{\textbf{C1}}
& \multicolumn{2}{c}{\textbf{C2}}
& \multicolumn{2}{c}{\textbf{C3}}
& \multicolumn{2}{c}{\textbf{C4}}
& \multicolumn{2}{c}{\textbf{All}} \\
\cmidrule(lr){2-3}
\cmidrule(lr){4-5}
\cmidrule(lr){6-7}
\cmidrule(lr){8-9}
\cmidrule(lr){10-11}

\textbf{Method}
& Gap & Col.\%
& Gap & Col.\%
& Gap & Col.\%
& Gap & Col.\%
& Gap & Col.\% \\
\midrule

RCPI
& \textbf{-0.15} & 0.00
& \textbf{0.16} & 0.00
& \textbf{0.63} & 0.00
& \textbf{1.56} & 0.02
& \textbf{0.55} & 0.01 \\

FVI
& 11.25 & 0.00
& 18.06 & 0.00
& 23.30 & 0.02
& 31.29 & 0.38
& 20.98 & 0.10 \\

DQN
& 7.24 & 0.00
& 11.11 & 0.16
& 19.98 & 0.36
& 35.26 & 1.06
& 18.40 & 0.40 \\

D-DDPG
& 6.03 & 0.00
& 6.60 & 0.00
& 9.96 & 0.12
& 17.09 & 0.40
& 9.92 & 0.13 \\

\bottomrule
\end{tabular}
\end{table*} 
Overall, RCPI achieves the smallest gap and the lowest collision rate among the four methods. D-DDPG ranks second with a smaller gap than both FVI and DQN. FVI and DQN have similar overall gaps, but DQN has the highest collision rate among the four methods. As the complexity increases from C1 to C4, both the gap and the collision rate generally increase. RCPI and D-DDPG maintain relatively stable gaps from C1 to C3 and their gaps increase in C4. The gap of FVI increases approximately linearly with the number of conflicting vehicles, whereas the gap of DQN increases at an accelerating rate. Although DQN has a smaller gap than FVI in C1--C3, it exceeds FVI in C4. Measured by the increase in the gap from C1 to C4, RCPI is the least sensitive to layout complexity (1.71), followed by D-DDPG (11.06), FVI (20.04), and DQN (28.02).

The ordering by collision rate does not coincide with the ordering by gap. D-DDPG has a smaller gap than FVI but a slightly higher collision rate. The reason is that collisions are rare in the test set. Over the five seeds, the learned policies are evaluated on $20{,}000$ trajectories in total, of which RCPI, FVI, D-DDPG, and DQN end in a collision in about 1, 20, 26, and 79 trajectories, respectively. The influence of these collisions is diluted when the discounted reward is averaged over all trajectories. Therefore, a smaller gap does not imply a lower collision rate.

RCPI differs from the other three methods in how it computes its target. RCPI obtains $Q^{\pi_k}(s,a)$ by summing the discounted rewards along a trajectory. Since the environment is deterministic, a single trajectory yields the exact value, and the target action contains no value-estimation error. Instead, FVI, DQN, and D-DDPG replace the true value of the next state with the prediction of a network. Any error of these networks therefore enters the target. Since the network is then updated toward this target and used to compute the next one, the error introduced at each update is carried forward and accumulates over training. This accumulation has been analyzed for
FVI~\cite{munos2008finite}, DQN~\cite{inproceedings}, and
actor-critic methods such as DDPG~\cite{fujimoto}.

\subsubsection{Training Stability}
\label{sec:stability}
\begin{figure}[t]
  \centering
  \includegraphics[width=\columnwidth]{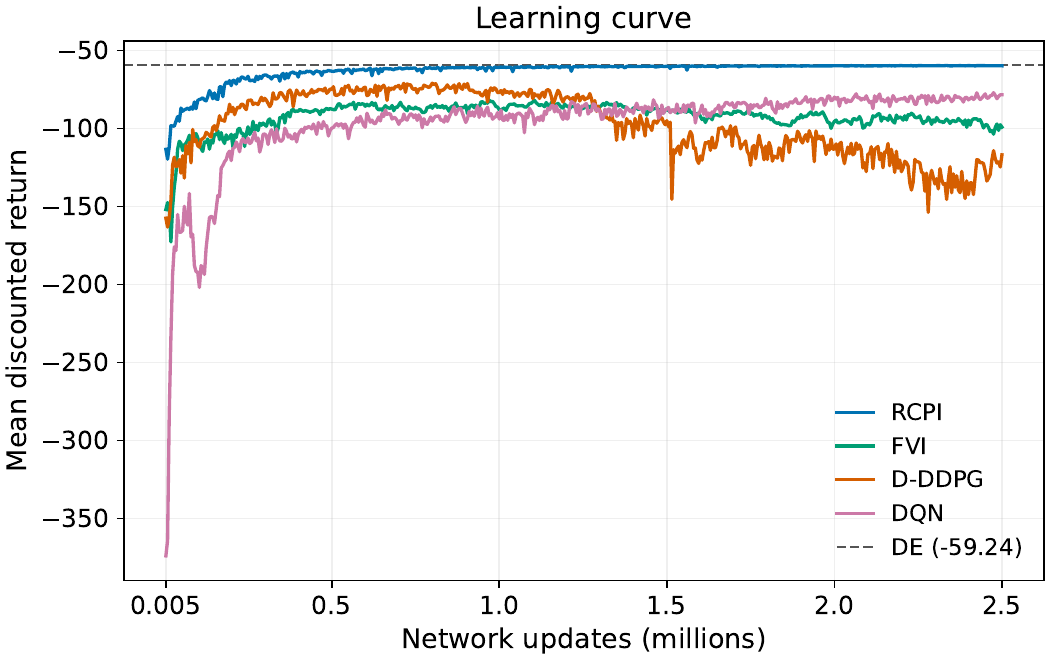}
  \caption{Discounted reward of the four methods versus the number of network updates.}
  \label{fig:learning_curve}
\end{figure}

Fig.~\ref{fig:learning_curve} and Table~\ref{tab:drop} present the training stability of each method.

\begin{table}[t]
  \centering
  \caption{Drop from the peak discounted reward to the discounted reward at the end of training. Mean $\pm$ standard deviation over five seeds.}
  \label{tab:drop}
  \begin{tabular}{@{}lc@{}}
    \toprule
    Method & Drop \\
    \midrule
    FVI    & $0.356 \pm 0.169$ \\
    RCPI   & $0.004 \pm 0.001$ \\
    DQN    & $0.105 \pm 0.011$ \\
    D-DDPG & $0.946 \pm 0.496$ \\
    \bottomrule
  \end{tabular}
\end{table}

From Fig.~\ref{fig:learning_curve}, we can see that RCPI improves and approaches the DE reference within about $0.5\times10^{6}$ updates. It then maintains this level until the end of training. From Table~\ref{tab:drop}, we can see that RCPI has the smallest drop among the four methods, which supports our observation. D-DDPG reaches the second-highest discounted reward at about $0.5\times10^{6}$ updates and starts to decline after about $1.3\times10^{6}$ updates. In the second half of training, D-DDPG shows large oscillations. Its drop of $0.946$ is the largest among the four methods. D-DDPG therefore has the least stable training process. DQN starts from the lowest discounted reward and oscillates strongly in the early updates. It then improves slowly but continuously. DQN has the second-smallest drop. FVI improves quickly in the early stage and then plateaus at a level clearly below the reference. Although the mean curve of FVI appears flat, its drop shows that each seed declines noticeably after reaching its peak.

As discussed above, RCPI is updated toward a target action that contains no value-estimation error. Its updates are correct from the start of training and do not rely on accurate value estimates of the next states. This allows RCPI to reach the reference level in fewer updates than the other methods. In contrast, the other three methods compute their targets from the estimates of their own networks, which are inaccurate in the early stage of training. They therefore need more updates for these estimates to become accurate before their targets become reliable. In D-DDPG, the critic is trained on the actions selected by the actor, and the actor is trained to maximize the output of the critic. If the critic estimates the value of an action inaccurately, the actor moves toward that action, and the critic is then trained on it. The errors of the two networks can therefore reinforce each other instead of being corrected. This may explain the large oscillations and the decline of D-DDPG in the later stage of training.

\subsubsection{Training and Decision Cost}
\label{sec:cost}
\begin{figure}[t]
  \centering
  \includegraphics[width=\columnwidth]{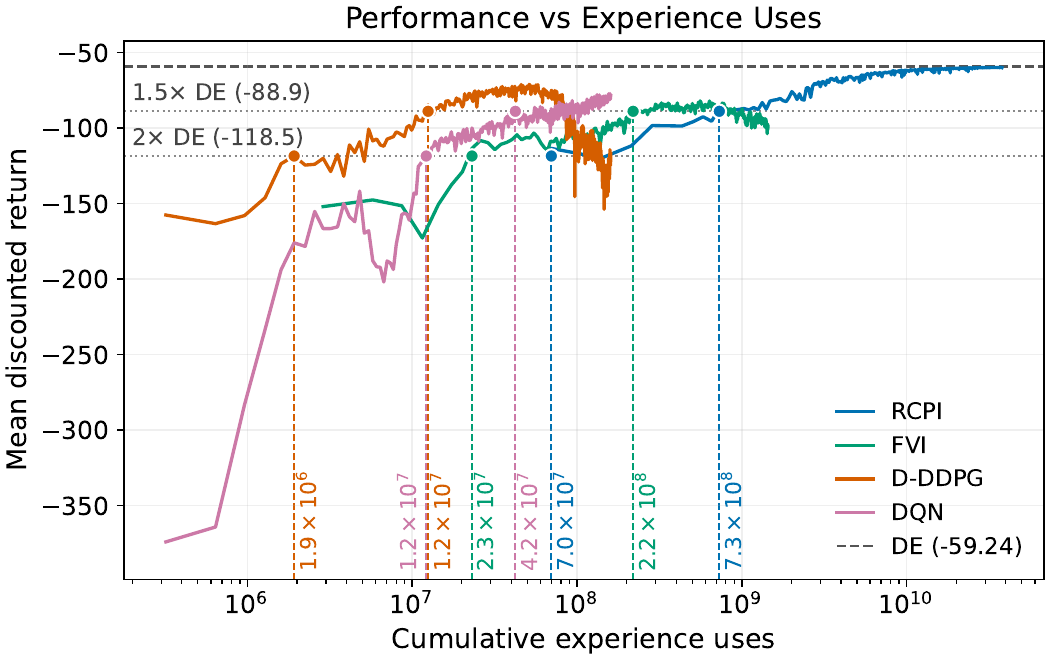}
  \caption{Discounted reward of the four methods versus the number of experiences used (log scale). The dashed lines mark $2\bar{V}^{\mathrm{DE}}$ and $1.5\bar{V}^{\mathrm{DE}}$.}
  \label{fig:experience_curve}
\end{figure}

Fig.~\ref{fig:experience_curve} shows the discounted reward of each method versus the number of experiences used. The dashed lines mark the thresholds $2\bar{V}^{\mathrm{DE}}$ and $1.5\bar{V}^{\mathrm{DE}}$. Table~\ref{tab:cost} reports the number of experiences used to reach the peak and the total number used in training.

\begin{table}[t]
\centering
\caption{Experience usage and decision time of each method. Mean $\pm$ SD over five seeds.}
\label{tab:cost}
\small
\setlength{\tabcolsep}{4pt}
\renewcommand{\arraystretch}{1.10}
\begin{tabular}{@{}lccc@{}}
\toprule
& \multicolumn{2}{c}{\textbf{Experiences}} & \textbf{Decision} \\
\cmidrule(lr){2-3}
\textbf{Method} & \textbf{Peak} & \textbf{Total} & \textbf{time (ms)} \\
\midrule
RCPI   & $(3.62 \pm 0.11)\times10^{10}$ & $3.79\times10^{10}$ & $0.051 \pm 0.003$ \\
FVI    & $(4.46 \pm 1.53)\times10^{8}$  & $1.44\times10^{9}$  & $0.227 \pm 0.007$ \\
DQN    & $(1.54 \pm 0.04)\times10^{8}$  & $1.60\times10^{8}$  & $0.048 \pm 0.003$ \\
D-DDPG & $(5.47 \pm 1.68)\times10^{7}$  & $1.60\times10^{8}$  & $0.050 \pm 0.003$ \\
\midrule
DE     & --- & --- & $370{,}416 \pm 9{,}326$ \\
\bottomrule
\end{tabular}
\end{table}

Fig.~\ref{fig:experience_curve} and Table~\ref{tab:cost} show that RCPI, which has the smallest gap, is the least data-efficient method. D-DDPG is the most data-efficient method, reaching each threshold with the fewest experiences, followed by DQN and FVI. RCPI requires the most experiences at every threshold. To reach its peak, RCPI uses $3.62\times10^{10}$ experiences. This is about 660 times as many as D-DDPG.

This order is consistent with the number of experiences required to compute one target. Although DQN and D-DDPG both require one experience per target, D-DDPG reaches each threshold with fewer experiences than DQN. A possible reason is how an experience affects the policy. In DQN, an experience $(s,a,r,s')$ updates only the estimate $Q_{\psi}(s,a)$ of the action $a$ it contains. Since the policy~\eqref{eq:dqn_policy} selects the action with the largest Q-value, the policy at $s$ changes only when this update is large enough to change which action has the largest value. In D-DDPG, the same experience updates both the critic and the actor. Each experience directly improves the policy. Thus D-DDPG is more data-efficient than DQN.
FVI is less data-efficient than DQN and D-DDPG because, to compute each target~\eqref{eq:ov_fvi_target} in our simulation, it needs to enumerate all actions in the action space, which requires $|\mathcal{A}|$ experiences. On the other hand, RCPI not only enumerates all actions but also simulates a trajectory for each action until termination, which requires $|\mathcal{A}|H$ experiences.

Table~\ref{tab:cost} also shows the decision time, measured according to the procedure in Section~\ref{sec:eval_setup}. The time is measured on a single core of an AMD EPYC 9454 CPU in the Owl cluster of Virginia Tech Advanced Research Computing (ARC). All four methods require much less decision time than DE. Among the four methods, RCPI, DQN, and D-DDPG have similar decision times. The training outcomes of RCPI and D-DDPG are policy networks that output the action directly. Although the training outcome of DQN is an action-value network, its action is obtained by taking the action with the largest value according to~\eqref{eq:dqn_policy}. In contrast, the training outcome of FVI is a state-value network. In our simulation, to select an action according to~\eqref{eq:fvi_policy}, FVI needs to enumerate all actions, obtain the reward and next state of each action, and evaluate the value network at each next state. FVI therefore requires a longer decision time than the other three methods.

\subsection{Method Selection}
\label{sec:selection}
The results in Section~\ref{sec:results} show that no single method is the best in all aspects. We therefore suggest selecting a method according to its features as follows.

RCPI achieves the smallest gap, the lowest collision rate, the most stable training, and the least sensitivity to layout complexity. Its limitation is that it requires a model of the environment during training and the largest number of experiences. RCPI is suggested when a model is available and the budget for training is not a constraint.

D-DDPG ranks second in optimality and is the most data-efficient method. It is model-free, so it can be trained by interacting with a simulator that cannot be queried at an arbitrary state. Its limitation is the least stable training. D-DDPG is suggested when no reliable model is available or the budget for training is limited. It is not suitable when the policy is updated with newly collected data during deployment, since the policy in use can degrade unpredictably~\cite{chan2020}.

DQN is also model-free and is the most stable among the model-free methods. However, it has the highest collision rate and is the most sensitive to layout complexity. DQN is suggested when no reliable model is available and stable training is required.

FVI does not achieve the best result in any single aspect. It requires the model not only during training but also at every decision according to~\eqref{eq:fvi_policy}, and its decision time therefore grows with the cost of querying the model. On the other hand, it requires far fewer experiences than RCPI, its training is more stable than that of D-DDPG, and its collision rate is lower than that of DQN. FVI is suggested as a compromise when a model is available at both training and deployment but the budget for training does not allow RCPI.

Since none of the four methods is collision-free, a safety mechanism is recommended at deployment regardless of the method selected, such as a short-horizon collision check~\cite{mokhtari2021}.

\section{Conclusions}
\label{sec:conclu}
RL methods for AV decision-making are rarely compared under identical conditions. This paper compares FVI, RCPI, DQN, and D-DDPG on the same decision-making problem at an unsignalized intersection in terms of optimality and safety, training stability, and training and decision cost. RCPI is the closest to the reference and the most stable method because it computes its target from exact trajectories, but it also requires the most experiences. D-DDPG ranks second in optimality and is the most data-efficient method, but its training is the least stable because the errors of its actor and critic can reinforce each other. DQN is the most stable model-free method but has the highest collision rate. FVI does not achieve the best result in any single aspect, but balances training cost, stability, and safety at the price of requiring the model at every decision. These features suggest selecting a method according to whether a model is available, the budget for training, and whether stable training is required, as discussed in Section~\ref{sec:selection}. The comparison is conducted in a deterministic environment with the exact model. Extending it to stochastic traffic and approximate models is left for future work.

\bibliographystyle{IEEEtran}
\bibliography{Reference}

\end{document}